\documentclass[10pt, a4paper]{article}

\usepackage[]{lrec-coling2024} 
\usepackage{xcolor}
\usepackage{hyperref}
 \definecolor{darkblue}{rgb}{0, 0, 0.5}
  \hypersetup{colorlinks=true, citecolor=darkblue, linkcolor=darkblue, urlcolor=darkblue}

\usepackage{xstring}

\usepackage{color}

\usepackage{color, colortbl}
\definecolor{LightCyan}{rgb}{0.88,1,1}
\definecolor{LightRed}{rgb}{1,0.8,0.7961}
\definecolor{LightGreen}{rgb}{0.8,1,0.7961}

\usepackage{graphicx}
\usepackage{amsmath}
\usepackage{hyperref}
\usepackage{tabularx}
\newcolumntype{C}{>{\centering\arraybackslash}X}

\usepackage{booktabs}
\usepackage{multirow, multicol}
\usepackage{comment}
\usepackage{float}
\restylefloat{table}
\usepackage{url}

\usepackage{algorithm}
\usepackage{algpseudocode}
\usepackage{arydshln}
\usepackage{array}
\usepackage{makecell}
\usepackage{multirow}
\usepackage{enumitem}

\usepackage{caption}

\usepackage{arydshln}

\def\alt{{\textsc{\textit{Alternate}}}}
\def\comb{{\textsc{\textit{Combine}}}}

\title{
Dynamic Reward Adjustment in Multi-Reward Reinforcement Learning for Counselor Reflection Generation
}

\name{Do June Min\textsuperscript{1}, 
        Ver\'{o}nica P\'{e}rez-Rosas\textsuperscript{1}, 
        Kenneth Resnicow\textsuperscript{2},
        Rada Mihalcea\textsuperscript{1}
}

\address{
       \textsuperscript{1}Department of Electrical Engineering and Computer Science,
       \textsuperscript{2}School of Public Health\\
      University of Michigan,
      Ann Arbor, MI, USA 
      \\
      \{dojmin, vrncapr, kresnic, mihalcea\}@umich.edu
}

\abstract{
In this paper, we study the problem of multi-reward reinforcement learning to jointly optimize for multiple text qualities for natural language generation. 
We focus on the task of counselor reflection generation, where we optimize the generators to simultaneously improve the fluency, coherence, and reflection quality of generated counselor responses.
We introduce two novel bandit methods, {\sc DynaOpt} and {\sc C-DynaOpt}, which rely on the broad strategy of combining rewards into a single value and optimizing them simultaneously. 
Specifically, we employ non-contextual and contextual multi-arm bandits to dynamically adjust multiple reward weights during training. 
Through automatic and manual evaluations, we show that our proposed techniques, {\sc DynaOpt} and {\sc C-DynaOpt}, outperform existing naive and bandit baselines, demonstrating their potential for enhancing language models.
\\ \newline \Keywords{multi-reward optimization, multi-armed bandits, reinforcement learning, linguistic rewards, policy optimization, reflection generation} }

\begin{document}
\maketitleabstract

\section{Introduction}
The field of natural language processing (NLP) has witnessed remarkable advancements in recent years, with reinforcement learning (RL) emerging as a powerful approach for optimizing language models \cite{li-etal-2016-deep, snell2023_ilql, Ramamurthy2022IsRL}. 
This paradigm has enabled practitioners to train models to align with diverse text properties and constraints, such as safety, helpfulness, or harmlessness \cite{Touvron2023Llama2O, bai2022training}. 
Central to this progress is the optimization of linguistic and behavioral constraints, which serve as guiding signals during the RL training phase, shaping the model's behavior toward desired objectives. 
However, as NLP tasks grow in complexity and diversity, a critical challenge arises when multiple linguistic rewards must be integrated into the training process. 

\begin{figure}[!htbp]
    \centering
    \includegraphics[width=0.8\linewidth,keepaspectratio]{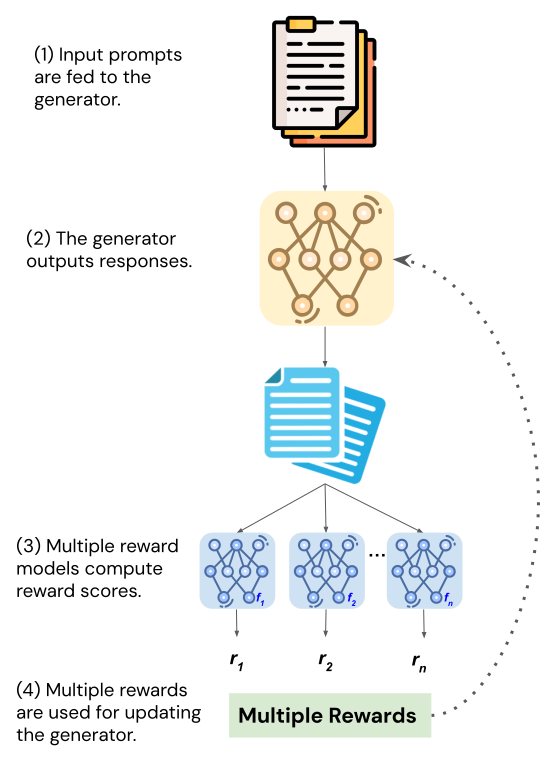}
    \caption{
    Workflow of RL training of language models with multiple rewards.
    The main question is how to simultaneously use the multiple rewards 
    to update the LM (Step 4).
    }
    \label{fig:multireward_optimization} 
\end{figure}

Researchers have explored different strategies to tackle the challenge of incorporating multiple rewards into the optimization of language models. 
Two prominent classes of methods have emerged in this context: (1) alternating between optimizing individual metrics at different points in the training process ({\alt}) \cite{pasunuru-bansal-2018-multi, zhou-etal-2019-building}, and (2) optimizing language models by simultaneously considering multiple metrics and combining their associated rewards into a unified objective ({\comb}) \cite{Sharma2021TowardsFE, yadav-etal-2021-reinforcement, deng-etal-2022-rlprompt}. 
The first approach, {\alt}, involves training language models by focusing on individual metrics at different stages of the training process, which can help address the challenge of incorporating multiple rewards by prioritizing each metric separately. 
The second approach, {\comb}, aims to optimize language models by simultaneously combining multiple metrics into a single unified objective, hence offering a more integrated approach when optimizing for various criteria. 

Previous methodologies employed for these approaches have predominantly relied on static configurations, wherein the alternating order or the combining ratio remains fixed and unchanging throughout the training process. 
To address this limitation, \citet{pasunuru-etal-2020-dorb} introduced the DORB extension within the {\alt} class of methods, which harnesses multi-armed bandit (MAB) algorithms to dynamically select the reward function to optimize at each stage of training. However, notably absent from their work is an exploration of how MABs can be leveraged to enhance and adapt the {\comb} class of multi-reward optimization methods, where rewards are jointly considered.

In this paper, we extend the {\alt} approach for multi-reward optimization by incorporating the dynamic control and adjustment of the mixing ratio of multiple rewards using MABs. 
Furthermore, we use contextual multi-armed bandits to address the absence of contextual information that could further aid in the optimization process. 
We evaluate our novel approaches against various baseline methods from both the {\alt} and {\comb} classes on counselor reflection generation, based on Motivational Interviewing counseling exchanges.  
Through our experiments, 
(1) we find that previous naive and bandit-based multi-reward optimization methods fall short of consistently improving training reward metrics, and 
(2) we show that our proposed methods, {\sc DynaOpt} and {\sc C-DynaOpt}, offer a comparative advantage in optimizing multiple rewards in the counselor response generation task, as shown in both automated and human evaluations.

We release our code at \url{https://github.com/michiganNLP/dynaopt}.

\section{Related Work}

Our work relates to several main research areas at the intersection of Machine Learning and NLP. 

\paragraph{Reinforcement Learning.}

RL has been successfully used to improve various NLP systems, including task-oriented dialogue systems, news article summarizers, and empathetic response generators~\cite{singh2022_optimizing, laban-etal-2021-keep, Sharma2021TowardsFE}.
These systems have applied various RL techniques to go beyond supervised learning with ground truth data by implementing reward models that provide learning signals to steer the behavior of language models
\cite{Ramamurthy2022IsRL}. RL strategies for this purpose include proximal policy optimization (PPO) \cite{SchulmanWDRK17}, self-critical sequence training (SCST) \cite{rennie2017}, implicit language Q-learning \cite{snell2023_ilql}, or Quark \cite{lu2022_quark} to name a few.
In our work, we chose the $k$-SCST algorithm for its simplicity and efficiency \cite{laban-etal-2021-keep}, but our approach is flexible enough to be used in tandem with RL optimization techniques.

\paragraph{Multi-reward Optimization.}
The inherent complexity of NLP tasks has motivated the use of multi-reward optimization in NLP methods~\cite{garbacea2022constrained, pmlr-v202-dann23a}. 
Particularly, in cases where 
defining a desired behavior for language models requires using multiple reward metrics, as single metrics often fall short of capturing the intricacies of model performance. 
Some examples can be found in \cite{li-etal-2016-diversity} which uses answering, coherence, and information flow, as multiple rewards for training dialogue agents, or in \citet{bai2022training, Touvron2023Llama2O}, where safety and helpfulness preference models serve as guiding rewards for the reinforcement learning with human feedback (RLHF) strategy.
Despite the widespread use of multiple rewards, the challenge of effectively combining them has received comparatively less attention in the literature. 
Notably, \citet{pasunuru-etal-2020-dorb} 
tackle this issue by dynamically selecting one reward for optimization during training. 

\begin{figure*}[t]
    \centering
    \includegraphics[width=1.0\linewidth,keepaspectratio]{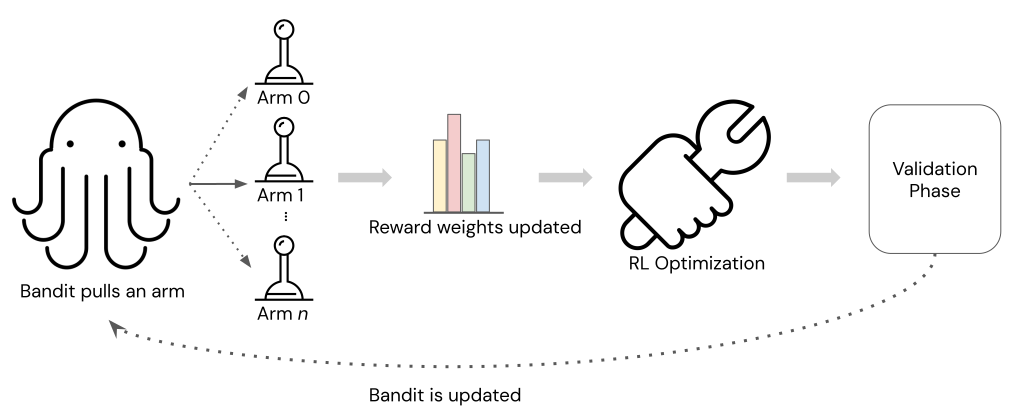}
    \caption{
    An overview of the {\sc DynaOpt} workflow.
    At each bandit step, the bandit pulls an arm, which updates the reward weights.
    Then, the RL optimization phase uses the summed weights and updates the language model.
    In the bandit update phase, the LM generations are scored by the reward models, and the
    scores are used to update the bandit model.
    }
    \label{fig:dynaopt} 
\end{figure*}

\paragraph{Multi-armed Bandits. }
MABs offer a framework for dynamically selecting among multiple actions and offer a way to navigate the exploration-exploitation trade-off \cite{AUDIBERT20091876, auer2002, auer2002_2, bubeck2012, burtini2015, burtini2015_2}.
This makes them suitable in NLP applications where dynamic control over multiple parameters, or input, selection is preferred over static control \cite{sokolov-etal-2016-learning}.
MABs have been successfully used for tasks such as news article recommendation \cite{li2010}, data selection in neural machine translation
\cite{kreutzer-etal-2021-bandits-dont}, model selection from a pool of multiple NLP systems \cite{Haffari2017EfficientBO}, and crowdsourced worker selection for annotation \cite{Wang2023CostefficientCF}.
In this work, we apply  MABs to the problem of multi-reward optimization.


\paragraph{NLP and Psychotherapy. }
Through our application testbed, our research is also closely connected to recent developments in NLP designed to support counselors in their practice and ongoing training within the counseling domain. 
Reflection, a critical element in counseling strategies like Motivational Interviewing, has been the subject of previous investigations to evaluate counseling.
For instance, practitioners have used the frequency and quality of reflections as a proxy for the quality of overall counseling \cite{flemotomosDBLP:journals/corr/abs-2102-11573, ardulov2022}. 
Furthermore, research has been conducted on the generation of reflections, where retrieval of human-written examples or relevant knowledge has been found to improve reflection generation performance \cite{shen-etal-2020-counseling, shen-etal-2022-knowledge, welivita-pu-2023-boosting}. 
In our work, we focus on exploring and comparing different multi-reward optimization techniques for RL training of counselor reflection generation models.

\section{{\sc DynaOpt}: Dynamically Adjusting Rewards for Multiple Rewards Reinforcement Learning}

%

We introduce {\sc DynaOpt}, a bandit method that enables {\comb} multi-reward optimization inspired by the Dynamically Optimizing Multiple Rewards with Bandits (DORB) framework~\cite{pasunuru-etal-2020-dorb}. 
{\sc DynaOpt} leverages the bandit framework to dynamically adjust the weights assigned to multiple rewards as shown in Figure~\ref{fig:dynaopt}.

\subsection{Multi-reward Optimization with Multi-armed Bandits}
The DORB framework 
employs the Exponential-weight algorithm for Exploration and Exploitation (Exp3) algorithm, which tackles the adversarial bandit problem \cite{auer2002, auer2002_2}, to dynamically select from a pool of reward functions during training stages. 
However, it only uses the {\alt} paradigm during the selection process. 
In contrast, {\sc DynaOpt} enables the use of the {\comb} strategy. 
When the Exp3 bandit chooses a reward, its corresponding weight is incremented.
In addition, while the DORB framework always chooses one reward at any given stage of training, {\sc DynaOpt} allows for a ``Do Nothing" option, which does not update the reward weights. Below, we describe the different steps followed by {\sc DynaOpt} to enable these strategies. The pseudocode for {\sc DynaOpt} is shown in Algorithm~\ref{alg:dynaopt}.

\paragraph{Choosing an Action.}
Given a reward function $f_i$ with $i \in {1,2,..,N}$ where $N$ is the number of rewards, the probability of selecting $f_i$ at time $t$ is given as:


\begin{equation}
\begin{aligned}
p_t(i) = (1 - \gamma) \frac{{ a_{t,i}}}{{\sum_{j=1}^{N} a_{t,j}}}  + \frac{\gamma}{N+1}
\end{aligned}
\label{eqn:exp3_select_arm}
\end{equation}
where $N+1$ indexes the choice of not updating reward weights, $\gamma$ is a mixing parameter for smoothing the probability with a uniform probability over the rewards. 
Importantly, the arm weights $a_{t,j}$ are distinct from the reward weights ($w_{t,j}$) themselves, and are only used to sample an action, which then updates the weights.

\paragraph{Updating Reward Weights. }
Our key idea is that given a bandit's reward selection, we increment its weight instead of optimizing for that reward only.
Specifically, when a bandit $B$ chooses reward $i$ at time $t$,
we adjust the weight of the reward function $W_{t+1,i}$ (Lines 5 \& 17, Algorithm~\ref{alg:dynaopt}), using Equations~\ref{eqn:exp3_update_1},\ref{eqn:exp3_update_2}, with $r^B_{t}$ set as $1$:
\begin{equation}
\begin{aligned}
\hat{r}^{B^W}_{t,j} = \begin{cases}
    \frac{r^{B^W}_{t}}{p_t(i)} & \text{if } j = i \\
    0 & \text{otherwise}
\end{cases}
\end{aligned}
\label{eqn:exp3_update_1}
\end{equation}
\begin{equation}
\begin{aligned}
w_{t+1,i} = w_{t,i} \exp\left(\frac{\gamma {\hat{r}^{B^W}}_{t,i}}{K}\right)
\end{aligned}
\label{eqn:exp3_update_2}
\end{equation}
Here, we reuse the Exp3 method of updating the bandit arm weights to update our reward weights.
The final reward weights $W$ can be obtained by normalizing over $w_{t+1,i}$.
Moreover, we introduce an additional weight $W_{N+1}$, with the $N+1$st arm representing the "Do Nothing" action, with 
$W_{t+1} = W_t$.

Although we reuse the weight update equation of the Exp3 algorithm,
we note that different weight adjustment schemes can be employed in {\sc DynaOpt}, allowing flexibility in adapting to various reward optimization scenarios.

\paragraph{Bandit Reward Computation.}
After optimizing the LM with the updated weights $W_t$, a reward must be computed to update the bandit $B$.
This can be obtained by measuring the performance of the updated LM over a validation set.
Departing from DORB, 
we use a different reward computation function for updating bandits (see line 4, Algorithm~\ref{alg:dynaopt}).
Instead of using scaled rewards, we define the bandit reward as the sum of the average improvement of reward $i$:

\begin{equation}
\begin{aligned}
\hat{r}^t &= \sum_i {r^t}_i\\
{r^t}_i = \text{Mean}&(R_{t,i}) - \text{Mean}(R_{t-1,i})
\end{aligned}
\label{eqn:compute_bandit_reward}
\end{equation}
where $R_{t,i}$ = is the history of unscaled rewards for function $i$ at time $t$.

\paragraph{Updating the Bandit.}
Given a reward for the bandit, we update the bandit arm weights using the Exp3 algorithm \cite{auer2002, pasunuru-etal-2020-dorb} (Line 15, Algorithm~\ref{alg:dynaopt}).
Updating the bandit arm weights $a_{t,i}$ uses the same formulas used to update the reward weights (Equations~\ref{eqn:exp3_update_1},\ref{eqn:exp3_update_2}).

\algrenewcommand\algorithmicrequire{\textbf{Input:}}
\algrenewcommand\algorithmicensure{\textbf{Output:}}
\begin{algorithm}[t]
\caption{{\sc DynaOpt} Optimization}\label{alg:dynaopt}
\hrulefill
\small
\begin{algorithmic}[1]
\Require{\# of rewards $N$, \# of train steps $n_{\text{train}}$, \# of RL validation steps $\text{round}_{\text{bandit}}$ 
Initial policy $p_0$, Initial Distribution of reward weights, $W$ (uniform distribution over $N$)
}
\State Make a copy $p_\theta$ of initial policy $p_0$.
\State Initialize Exp3 bandit $B$ with $N+1$ arms.
\State Initialize weights $w_{0,i}$ over $i\in[1,2,\cdots,N+1]$ as uniform distribution.
\State $a$ $\leftarrow$ chooseArm($B$) \Comment Eqn~\ref{eqn:exp3_select_arm}
\State $W$ $\leftarrow$ UpdateRewardWeight($W, B_w, a, 1$) \Comment Eqns~\ref{eqn:exp3_update_1},\ref{eqn:exp3_update_2}
\State $i \leftarrow 0$
\While{$i < n_\text{train}$} 
\State train\_responses $\leftarrow$ Sample($p_\theta$, train\_data)
\State $r_\text{train}$ $\leftarrow$ ComputeReward(train\_responses, $W$)
\State Optimize $p_\theta$ with $R_\text{train}, p_0$ \Comment Eqn~\ref{eqn:rl_loss}
\If{i \% $\text{round}_{\text{bandit}} == 0$}
\State dev\_responses $\leftarrow$ Sample($p_\theta$, dev\_data)
\State $r_\text{bandit}$ $\leftarrow$ ComputeReward(dev\_responses, uniform weights)
\State $r \leftarrow$ ComputeBanditReward($r_\text{bandit}$) \Comment Eqn~\ref{eqn:compute_bandit_reward}
\State UpdateBandit($B$, $a$, $r$) \Comment Eqns~\ref{eqn:exp3_update_1},\ref{eqn:exp3_update_2}
\State $a$ $\leftarrow$ chooseArm($B$) 
\State $W$ $\leftarrow$ UpdateRewardWeight($W, B_w, a, 1$) \Comment Eqns~\ref{eqn:exp3_update_1},\ref{eqn:exp3_update_2} 
\EndIf
\State $i \leftarrow i + 1$
\EndWhile
\end{algorithmic}
\end{algorithm}

\subsection{{\sc C-DynaOpt}: Reward Update with Contextual Multi-armed Bandits.}
Contextual multi-armed bandits (CMABs) are a class of decision-making algorithms that incorporate contextual information to optimize action selection. 
In contrast to traditional MABs, which rely solely on historical performance, contextual MABs utilize additional context to make more informed and adaptive choices in various scenarios \cite{cortes2018, agarwal2014}.

We extend the {\sc DynaOpt} algorithm by using \textit{contextual} MABs
\cite{burtini2015, bietti2021} instead of non-contextual MABs.
We use Vowpal Wabbit's algorithm\footnote{Vowpal Wabbit \href{https://vowpalwabbit.org}{https://vowpalwabbit.org}} to replace the Exp3 bandit and provide the current reward weights $W_t$ and average RL dev set reward for each reward function as context to the bandit algorithm.



\section{Datasets and Task}

\subsection{Datasets}
We use two counselor reflection datasets in our experiments. Table~\ref{table:dataset_stats} shows statistics for these datasets. We combine and shuffle the datasets and use a split of 50\%/10\%/40\% for our train/dev/test split.

\paragraph{PAIR Dataset.}
PAIR is a compilation of interactions between clients and counselors consisting of single-turn exchanges. 
It was curated by \cite{min-etal-2022-pair}. 
The data collection process involved a combination of expert and crowdsource annotations. 
Expert annotations were employed for the reflection category, which demands proficiency in Motivational Interviewing, while crowdsource annotations were utilized for capturing non-reflections containing directive language.
Following the Motivational Interviewing Treatment Integrity (MITI) code established in Motivational Interviewing literature by \cite{moyers2016}, each counselor's response is categorized as Complex Reflection (CR), Simple Reflection (SR), or Non-Reflection (NR). 
Examples illustrating these categories can be found in the bottom three rows of Table~\ref{tab:model_generations}. 
Complex Reflections (CRs) are deemed as superior responses in comparison to Simple Reflections (SRs), which, in turn, are rated higher than Non-Reflections (NRs).

\paragraph{CounselChat and Reddit Dataset.}
The dataset compiled by \citet{welivita-pu-2023-boosting} comprises dialogues extracted from online peer support forums, such as CounselChat and Reddit, providing a valuable but challenging resource for training AI-driven chatbots aimed at addressing psychological distress. 
These dialogues feature a mixture of conforming and non-conforming responses, reflecting the diverse nature of peer support interactions. 
The authors also leverage the MITI code, a well-established behavioral coding scheme to categorize these response types. 
We further process to extract only client prompt and counselor responses where the counselor responses are either simple or complex reflections, based on the MITI coding of the utterances.

\begin{table}[!htbp]
\small
\centering
\resizebox{0.95\linewidth}{!}{%
\begin{tabular}{lcc}
\toprule
Statistics & PAIR  & \citet{welivita-pu-2023-boosting} \\ \midrule
\# of Exchange Pairs & 2544 & 1184 \\
Avg \# of Words & 32.39 & 36.92 \\
\# of Complex Reflection & 636 & 768\\
\# of Simple Reflection & 318 & 416 \\
\# of Non-Reflection & 1590 & 0 \\\bottomrule
\end{tabular}
}
\caption{Counselor reflection dataset statistics.
}
\label{table:dataset_stats}
\end{table}

\subsection{Counselor Reflection Generation}
Counselor reflection generation is the task of automatically generating empathetic responses that mirror and affirm a client's thoughts and feelings, fostering a therapeutic and collaborative dialogue. While multiple counseling strategies are available, we follow the Motivational Interviewing strategy. Specifically, we aim to generate counselor reflections (Complex and Simple) given client prompts about issues such as drug cessation, weight loss, or health problems. 


\paragraph{Reward Metrics.} We use three main metrics to measure the quality of generated counselor reflections in terms of counseling style, fluency and coherence. 

\begin{itemize}[leftmargin=*]
    \item Reflection Score \cite{min-etal-2022-pair}: This metric quantifies the quality and relevance of generated counselor reflections, ensuring that the model produces responses that are contextually appropriate and meaningful within a counseling context.
    It is computed by a RoBERTa scoring model that was pretrained using the PAIR dataset.
    We use the original weights trained by \citet{min-etal-2022-pair}.
    \item Fluency: Fluency is assessed to evaluate the smoothness and coherence of generated reflections, ensuring that they read naturally and coherently. We implement our fluency reward as the inverse of the perplexity of the generated responses, following \cite{Sharma2021TowardsFE}.
    \item Coherence: Coherence evaluates the logical flow and consistency of the generated counselor reflections. 
    We implement the coherence reward by training a RoBERTa classifier trained to detect coherent and incoherent client prompt and counselor response pairs, where incoherent pairs are created by matching prompts to randomly sampled responses \cite{Sharma2021TowardsFE}.
\end{itemize}

\paragraph{Evaluation Metrics.}
In our evaluation, we extend our assessment beyond the reward metrics and delve into two additional linguistic-based metrics.  
\begin{itemize}[leftmargin=*]
    \item Diversity (dist-2) \cite{li-etal-2016-diversity}: This metric gauges the linguistic diversity of the generated counselor reflections. It measures the variety and richness of language used in the model's responses.
    \item The Levenshtein edit rate (Edit Rate): This quantifies the extent to which the model successfully avoids verbatim repetition of client words. This aspect of the evaluation ensures that the generated counselor reflections strike a balance between maintaining consistency and avoiding excessive repetition, aligning with the principles of effective counseling within MI \cite{lord2014_mi}.
\end{itemize}

\paragraph{Human Evaluation.}
In addition to automated evaluation, we conducted human annotation of 100 randomly sampled generated reflections from four models (DORB, Uniform Weighted, {\sc DynaOpt}, {\sc C-DynaOpt}) to assess generation quality. 
Two motivational interviewing experts collaborated as consultants, rating generations on a 3-level scale. 
Then, the ratings were normalized to [0, 1].
Our guidelines for the human annotators are included below: 

\paragraph{Human Annotation Guidelines.}
\label{subsection:annotation_guideline}
The reflection ratings were based on the MITI coding system \cite{moyers2016}. The following guidelines were provided to the human annotators:\\


\noindent \underline{Reflection.}
To evaluate responses as either 0 (Non-Reflection), 1( Simple Reflection), or 2 (Complex Reflection), follow these guidelines:

\begin{itemize}

\item Non-Reflection (0):
    A response is considered a non-reflection when it does not engage with the client's input or the task at hand.
    It may be off-topic, irrelevant, or simply fail to address the client's query.

\item Simple Reflection (1):
    A response is categorized as a simple reflection when it acknowledges the client's input or question without adding substantial depth or insight.
    It might repeat or rephrase the client's words, showing understanding but not extending the conversation significantly.
    Simple reflections demonstrate basic engagement with the client's query.

\item  Complex Reflection (2):
    A response is identified as a complex reflection when it goes beyond mere acknowledgment and engages deeply with the client's input or question.
    It demonstrates an understanding of the client's thoughts, feelings, or concerns and provides a thoughtful, insightful, or elaborate response.
    Complex reflections contribute to the conversation by expanding upon the client's ideas or by offering new perspectives and information.
    
\end{itemize}
When evaluating responses, choose the most appropriate category (0, 1, or 2) based on these criteria. 
Keep in mind that responses may vary in complexity, and your judgment should be guided by the degree to which they reflect upon the client's prompt.\\

\noindent \underline{Coherence.} Rate the coherence of the counselor on a scale of 0 to 2 (0=not coherent at all, 1=somewhat coherent, 2=very coherent). 
Coherent counselor responses should effectively address the client's concerns and maintain a logical flow of conversation.\\

\noindent \underline{Fluency.} Assess the linguistic naturalness and smoothness of the counselor's responses. 
Responses are rated on a scale from 0 to 2, where 0 indicates responses that lack fluency, 1 signifies somewhat fluent responses, and 2 represents responses that are highly fluent and natural in their expression. 
Fluent counselor responses should convey information in a clear and easily understandable manner, ensuring effective communication with the client. 

\begin{table*}[!htbp]
    \centering
    \small
\resizebox{\textwidth}{!}{%
    \begin{tabular}{c|ccc||cc}
    \toprule
         Models&  Reflection ($\uparrow$) &  Fluency ($\uparrow$) &  Coherence ($\uparrow$) &  Edit Rate ($\uparrow$)  & Diversity-2 ($\uparrow$) \\
         \midrule
Round& -5.02\% & 11.36\% & 5.51\% & -8.75\% & -0.20\% \\
\rowcolor{LightGreen} Uniform Weighted& 4.48\%  & 8.13\%  & \textbf{5.36}\% & -6.28\% & \textbf{-0.23}\% \\
DORB \cite{pasunuru-etal-2020-dorb}& -3.03\% & 9.54\%  & 5.42\% & -7.00\% & -0.08\% \\
\rowcolor{LightGreen} {\sc DynaOpt}& \textbf{7.80}\%  & 7.03\%  & 5.02\% & \textbf{-4.90}\% & -0.63\% \\
\rowcolor{LightGreen} {\sc C-DynaOpt}& 6.14\%  & \textbf{8.73}\%  & 5.02\% & -5.75\% & -0.46\% \\
 \bottomrule
    \end{tabular}
}
    \caption{Automated evaluation results on the counselor reflection generation task. 
    We compute the average measurements of 5 different runs and report relative change over the Cross Entropy baseline. 
    \textcolor{green}{Green} indicates the model achieved improvement over all reward metrics.
    }
    \label{tab:reflection_automated}
\end{table*}
\section{Experiments}

\subsection{RL Algorithm}
In our experiments, we employ the $k$-Self-Critical Sequence Training ($k$-SCST) algorithm~\cite{laban-etal-2021-keep},  for its simplicity and effectiveness. 
In the $k$-SCST technique, $k$ $(\ge2)$ samples are generated, and then their rewards
$R^{S_1},\cdots, R^{S_k}$, alongside the average reward achieved by the samples, $\overline{R}^S$, which serves as the baseline.

In addition, we use a KL-divergence loss between the initial policy $p_0$ and trained policy $p_\theta$ to prevent the model from deviating from the original model and generating unnatural text \cite{Ramamurthy2022IsRL}.
Thus, our RL training objective is as follows:
\begin{equation}
\begin{aligned}
L_{RL} = \frac{1}{k}\sum_{j=1}^{k} [ (\overline{R}^S - {R}^{S_j}) 
\log p_\theta(\cdot| c)
\\- \beta 
\text{ KL}( p_\theta(\cdot | c) &\| p_{0}(\cdot | c) ) ]
\end{aligned}
\label{eqn:rl_loss}
\end{equation}
where
$c$ is the prompt, $\cdot$ is the generated sequence conditioned on $c$, $\beta$ is the KL divergence coefficient. 

\subsection{Models}
We evaluate our proposed approaches, {\sc DynaOpt} and {\sc C-DynaOpt}, alongside several baselines.

\begin{itemize}[leftmargin=*]
    \item Cross Entropy: This model is trained using standard supervised learning methods. 
    It serves as a warm-start model for all the RL-trained models, providing an initial reference point for comparison.
    \item Round:
    Within the {\alt} category, the Round baseline employs a fixed round-robin strategy, cyclically switching between reward functions. Each reward function is allocated a round size of $20$ steps. This approach represents a simplistic but systematic way of alternating between different rewards during training.

    \item Uniform Weighted: Falling under the {\comb} class, the Uniform Weighted baseline employs a straightforward uniform weighting scheme to average the reward metrics. 

    \item DORB: We implement the Single Multi-armed Bandit (SM) DORB method proposed in \citet{pasunuru-etal-2020-dorb}'s.
    We focus on the single bandit version of DORB since previous studies have shown 
on par or superior performance to more complex hierarchical models \cite{pasunuru-etal-2020-dorb}.
Moreover, we avoid the additional overhead required while evaluating and tuning the controller bandit.

\end{itemize}
\subsection{Experimental Setup}
We tune our hyperparameters 
on the validation. 
We train and test the pretrained \texttt{t5-base} model \cite{raffel2020} on Nvidia's GeForce GTX 2080 GPUs, and use a batch size of 10, which is also the $k$ parameter for the $k$-SCST algorithm.

We report the averaged results of 5 different runs for our automated evaluation.

\begin{figure*}[]
\begin{minipage}[b]{0.5 \linewidth}
\centering
\Huge
\resizebox{1.0\linewidth}{!}{%
\begin{tabular}{ll}\toprule
Client Prompt & \begin{tabular}[c]{@{}l@{}}
I’m a teenager. My entire family needs\\ 
family therapy, and more than likely individual \\
therapy. My parents refuse to take action, and \\
I'm tired of it. Is there any way I can get out \\
of this myself?
\end{tabular} \\ \midrule\midrule
\begin{tabular}[c]{@{}l@{}}Uniform Weighted  \end{tabular}
& \begin{tabular}[c]{@{}l@{}}
Your parents are telling you that family\\
therapy is needed for your entire family.\\
You're worried that as a teenager, it's \\
getting too overwhelming and overwhelming.
\end{tabular} \\ \midrule
\begin{tabular}[c]{@{}l@{}} DORB (\citeyear{pasunuru-etal-2020-dorb}) 
\end{tabular}& \begin{tabular}[c]{@{}l@{}}
Your parents are fed up with what they see \\
as a lack of action from your part of the\\
family.
\end{tabular} \\ \midrule
\begin{tabular}[c]{@{}l@{}}  {\sc DynaOpt} \end{tabular} & \begin{tabular}[c]{@{}l@{}}
You're questioning whether or not you\\
can get help for your entire family.
\end{tabular} \\ \midrule
\begin{tabular}[c]{@{}l@{}}  {\sc C-DynaOpt} \end{tabular} & \begin{tabular}[c]{@{}l@{}}
You're questioning if there's any way \\
you can get out of this situation.
\end{tabular} \\ \midrule\midrule
\begin{tabular}[c]{@{}l@{}} Complex Reflection \\ (Human) \end{tabular}& \begin{tabular}[c]{@{}l@{}}
You're frustrated that your parents are hesitant\\
to take action, and you're wondering if there's \\
a way for you to address these concerns on your\\
own.
\end{tabular} \\ \midrule
\begin{tabular}[c]{@{}l@{}} Simple Reflection \\ (Human) \end{tabular}& \begin{tabular}[c]{@{}l@{}}
You're tired of your parents not taking action.
\end{tabular} \\ \midrule
\begin{tabular}[c]{@{}l@{}} Non-Reflection \\ (Human) \end{tabular}& \begin{tabular}[c]{@{}l@{}}
Have you talked to your parents about how\\ you're feeling?
\end{tabular} \\\bottomrule
\end{tabular}
}
\captionof{figure}{Sample reflection generations of different models on the counselor reflection generation task.}
\label{tab:model_generations}
\end{minipage}\hfill
\begin{minipage}[b]{0.45\linewidth}
\resizebox{\textwidth}{!}{%
\begin{tabular}{c|ccccc}
\toprule
 & \begin{tabular}[c]{@{}c@{}}Uniform\\Weighted\end{tabular}   & \begin{tabular}[c]{@{}c@{}}DORB\\(\citeyear{pasunuru-etal-2020-dorb})\end{tabular}& {\sc DynaOpt} & {\sc C-DynaOpt} \\
\midrule
Reflection ($\uparrow$) &  69.44                     & 64.16                  & 73.50                     & 74.10                       \\
Fluency ($\uparrow$) &  45.65                     & 47.40                  & 46.27                     & 46.58                       \\
Coherence ($\uparrow$) & 86.49                     & 86.86                  & 86.37                     & 86.40                       \\
\hdashline
Edit Rate ($\uparrow$) & 83.79                     & 83.11                  & 85.52                     & 85.08                       \\
Diversity-2 ($\uparrow$) & 92.07                     & 92.35                  & 91.41                     & 91.80                \\
\bottomrule
\end{tabular}
}
\captionof{table}{
Automated evaluation results on the counselor reflection generation task (run seed $=x$).
}
\label{tab:automated_one_run_reflection}
\vspace{0.35in}
    \resizebox{\linewidth}{!}{%
        \begin{tabular}{c|cccc}
            \toprule
             & \begin{tabular}[c]{@{}c@{}}Uniform\\Weighted\end{tabular}  & \begin{tabular}[c]{@{}c@{}}DORB\\(\citeyear{pasunuru-etal-2020-dorb}) \end{tabular} & {\sc DynaOpt} & {\sc C-DynaOpt} \\
            \midrule
            Reflection & 28.29 & 25.30 & \textbf{32.10} & 29.93 \\
            Fluency & \textbf{60.31} & 55.85 & 59.38 & 58.91  \\
            Coherence & 62.48 & 62.79 & 63.62 & \textbf{63.68}  \\
            \bottomrule
        \end{tabular}
    }
    \captionof{table}{Human evaluation results on the counselor reflection dataset.}
    \label{tab:human_reflection_evaluation}
\end{minipage}
\end{figure*}
\section{Results and Analyses}

\subsection{Overall Results}
\paragraph{Not All Multi-reward Optimization Methods Are Effective for Counselor Reflection Generation.}
In our experiments,
methods within the {\comb} class exhibit superior performance compared to the {\alt}  methods (Table~\ref{tab:reflection_automated}). 
We note that {\comb} methods such as DORB or Round failed to improve over the Cross Entropy baselines in automated reflection evaluation.
This result is in contrast to \citet{pasunuru-etal-2020-dorb}'s experiments on data-to-text generation and question generation tasks, which showed that their {\alt}-based bandit method DORB was able to achieve improvements in overall metrics.
This stresses the absence of a universally optimal method for multi-reward optimization, highlighting the nuanced nature of reward combinations and the influence of task specifics on the efficacy of different approaches.

\paragraph{Comparative Advantage of Our  Methods.}
Our results show that {\sc DynaOpt} and {\sc C-DynaOpt} outperform not only the {\alt} methods but also the Uniform Weighted baseline in terms of both automatic and human reflection levels while achieving similar levels in other metrics (Tables~\ref{tab:reflection_automated},~\ref{tab:human_reflection_evaluation}).
Specifically, while the bandit-based DORB training leads to degraded performance over automated reflection, our methods show consistent improvement over all reward metrics. 


\begin{figure*}[]
    \centering
    \begin{minipage}{.475\textwidth}
        \centering
        \includegraphics[width=\textwidth]{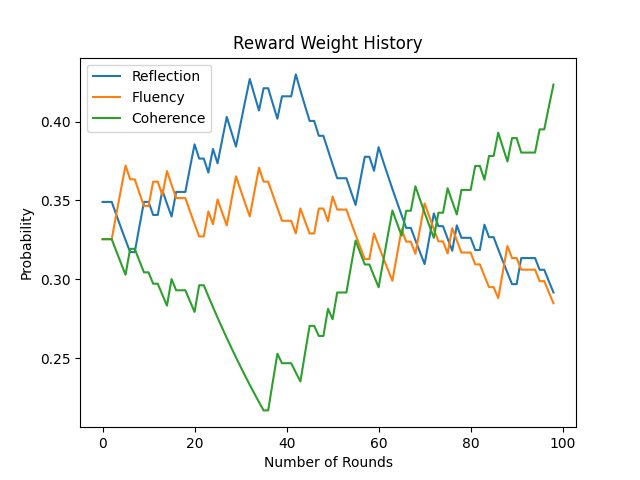}
        \caption{Reward weight trajectory of the {\sc DynaOpt} model on the counselor reflection generation task.}
        \label{fig:cdynaopt_pmf_history} 
    \end{minipage}%
    \hfill
    \begin{minipage}{0.475\textwidth}
        \centering
        \includegraphics[width=\textwidth]{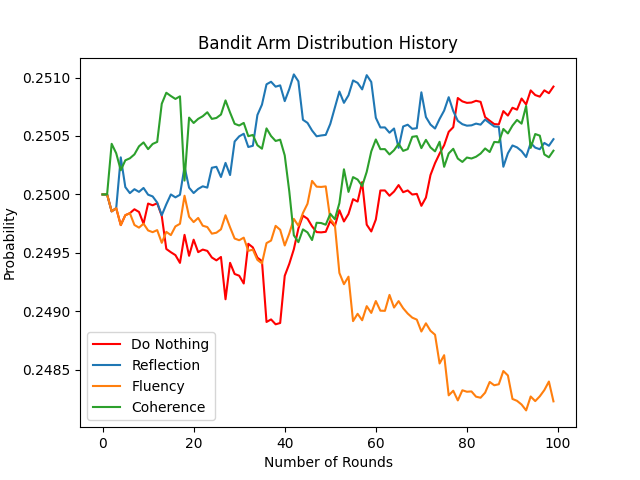}
        \caption{Bandit arm weight history of the {\sc DynaOpt} model on the counselor reflection generation task.
        }
        \label{fig:reward_change} 
    \end{minipage}
\end{figure*}

\subsection{Automated Evaluation}
In the Counselor Reflection Generation experiment (see Table~\ref{tab:reflection_automated}), we observe 
that the {\comb} methods over performed the {\alt} models. 
While the {\comb} and {\alt} models achieve similar levels of Fluency and Coherence, the {\comb} models exhibit notable improvements over the Cross Entropy baseline in terms of Reflection measurements. 
In contrast, the {\alt} models show degraded performance in the Reflection metric.
Interestingly, DORB achieves higher reflection levels compared to the Round approach. 

Furthermore, in contrast to the bandit-based models, the Round approach exhibits higher overall variance over random runs (reflection variance of $3.59$ vs $1.43$ \& $1.29$ of our methods),
indicating less stability in the training process. 
This further suggests that in certain settings, the bandit approach may contribute to a more stable and adaptable training process by dynamically adjusting reward weights and optimizing multiple rewards as training progresses. 

\subsection{Human Evaluation}
The results of our human evaluation are presented in Table~\ref{tab:human_reflection_evaluation}, with sample-generated reflections shown in Table~\ref{tab:model_generations}. 
In this evaluation, we compared the bandit-based approaches (DORB, {\sc DynaOpt}, {\sc C-DynaOpt}) with the Uniform Weighted model.
For human evaluation, we utilized the models trained in a single run (automated evaluations are shown in Table~\ref{tab:automated_one_run_reflection}). 
%

The evaluation results confirm the trends observed in our automated evaluation. 
Specifically, the {\comb} models outperform the Uniform Weighted model, which falls within the {\alt} class. 
Among the {\comb} models, our proposed approaches ({\sc DynaOpt} and {\sc C-DynaOpt})  outperform the Uniform Weighted baseline in achieved reflection.
We also see that in terms of human-evaluated fluency and coherence, our models slightly outperform DORB despite having lower automated results.
This could be attributed to the higher reflection levels contributing to the overall naturalness of the generated responses of {\alt} models.
Our human evaluation reaffirms the effectiveness of our bandit-based methods, particularly in terms of enhancing reflection quality in counselor responses.

\subsection{Bandit Visualization}

To understand the dynamics of bandit-based reward adjustment for {\sc DynaOpt}
we visualized the trajectory of reward weights over the RL development set throughout training.\footnote{For all our visualizations we use the same random seed run used for results reported in Tables~\ref{tab:automated_one_run_reflection}
 and ~\ref{tab:human_reflection_evaluation}} 
Notably, we observe that the relative importance of each reward dynamically changes over time, underscoring the adaptive nature of our bandit-based control of reward weights. 
This dynamic adjustment allows the model to optimize multiple rewards effectively as training progresses.

We also plot the history of the probability distribution of each arm during training in Figure~\ref{fig:reward_change}, where "Do Nothing" corresponds to the action of not updating the reward weight distribution.
We note that the trajectory of the reward weight history can be understood by tracking the evolution of bandit arm probabilities.
For example, the increase of coherence reward weight around round \#40 coincides with the corresponding increase of the coherence arm weight, the decline of reflection and fluency weights, as well as the rapid boost in the "Do Nothing" arm.

\section{Conclusion}

Our study addressed the problem of optimizing multiple linguistic rewards in reinforcement learning in the context of counselor reflection generation in motivational interviewing (MI). 
We explored two primary optimization strategies, the {\alt} and {\comb} approaches, and also presented bandit-augmented versions of the latter class.
Our two novel bandit methods, {\sc DynaOpt} and {\sc C-DynaOpt}, operate by dynamically adjusting reward weights during training using multi-armed bandits. 
Based on our empirical assessments, we observed that previous naive and bandit-based approaches to multi-reward optimization fail to improve response generations over the reward metrics.
In addition, our proposed techniques, {\sc DynaOpt} and {\sc C-DynaOpt}, outperform existing baselines in the counselor response generation task, demonstrating their potential for enhancing the RL step of training language models.

\section{Limitations}

There are limitations in our study that suggest directions for future investigation. 
First, we have yet to examine whether the trajectory of rewards during training influences holistic model behavior, especially when {\alt} and {\comb} models achieve similar performance metrics after optimization. 
In addition, our study's focus on moderate-scale language models overlooks the implications of applying our approach to larger models with billions of parameters, such as Llama 2 \cite{Touvron2023Llama2O}.
Also, although our method is independent of specific RL optimization algorithms, we have not conducted experiments with popular RL algorithms such as proximal policy optimization \cite{SchulmanWDRK17}. 
Addressing these limitations in future research will help provide a more comprehensive understanding of the applicability and efficacy of our approach in a wider range of contexts and settings.
We also emphasize that in this study our priority was to explore and compare different strategies for optimizing reflection generators with reinforcement learning, rather than creating state-of-the-art models.


\section{Ethical Considerations}
The datasets used in our study include motivational interviewing conversations between counselors and patients.
We ensured that the source datasets processed the dialogues so that personally identifiable information was redacted.
In addition, we stress that we do not advocate for the deployment of our models in clinical or mental health settings,
both because human understanding and communication are vital in these domains and the behavior of language models is not fully understood.
We recommend that current MI and counseling systems are best considered as tools that are best used for training and coaching learning practitioners.

\section*{Acknowledgements}

The authors would like to thank researchers and students from the University of Michigan School of Public Health, for their valuable feedback and participation in this project. This material is based in part upon work supported by
a National Science Foundation award (\#2306372). Any opinions, findings, conclusions, or recommendations expressed in this material are those of the authors and do not necessarily reflect the views of the NSF.

\section{Bibliographical References}\label{reference}


\bibliographystyle{lrec_natbib}
\bibliography{bib}



\appendix
\section{Appendix}
\label{sec:appendix}

\subsection{DORB Bandit Reward Computation}
Following \citet{graves17}, \citet{pasunuru-etal-2020-dorb} defines the bandit reward at time $t$ as a mean of scaled rewards:
\begin{equation}
    \hat{r}^t =
\begin{cases}
0 & \text{if } R_t < q^{{lo}}_t \\
1 & \text{if } R_t > q^{{hi}}_t \\
\frac{R_t - q^{{lo}}_t}{q^{{hi}}_t - q^{{lo}}_t} & \text{otherwise}
\end{cases}
\end{equation}
where $R_t$ = is the history of unscaled rewards at time $t$ and $q^{{lo}}_t, q^{{hi}}_t$ are the lower and upper quantiles of $R_t$.



\subsection{Experiment Hyperparameters}
We include the hyperparameter values we used in Table~\ref{table:experiment_parameters}.
We optimize the learning rate, the bandit coefficient $\gamma$, and the parameters of the contextual bandit exploration's online cover by conducting a grid search based on the criterion of average reward maximization.
\begin{table}[!htbp]
\centering
\resizebox{\linewidth}{!}{%
\begin{tabular}{lc}
\toprule
\multicolumn{2}{c}{Supervised Learning (Cross Entropy model)} \\
\midrule
Language Model & \texttt{t5-base} \\
Training epochs & 5\\
Learning Rate & 1e-4 \\
\midrule
\multicolumn{2}{c}{Reinforcement Learning ($k$-SCST)} \\
\midrule
Language Model & \begin{tabular}[c]{@{}c@{}}\texttt{t5-base}\end{tabular}\\
Learning Rate & 1e-4 \\
Sampling Temperature & 1.0 \\
Testing Temperature & 0.5 \\
$k$ & 10 \\
KL weight $\beta$ & 0.05 \\
$n_{\text{train}}$ & 1000 \\
$n_{\text{bandit}}$ & 200 \\
$\text{round}_{\text{bandit}}$ & 10 \\
bandit coefficient $\gamma$ & 0.07 \\
Bandit History Size $H$ & 200 \\
Contextual Bandit Exploration & Online Cover $=3$\\
\bottomrule
\end{tabular}
}
\caption{Experiment models \& parameters.}
\label{table:experiment_parameters}
\end{table}

\section{Evaluation Results in Absolute Number}
We include the automated evaluation results in absolute values and also include the standard deviation over the five different runs (Table~\ref{tab:reflection_automated_absolute}).

\begin{table*}[h!]
    \centering
\resizebox{\textwidth}{!}{%
    \begin{tabular}{c|ccc||cc}
    \toprule
         Models&  Reflection ($\uparrow$) &  Fluency ($\uparrow$) &  Coherence ($\uparrow$) &  Edit Rate ($\uparrow$)  & Diversity-2 ($\uparrow$) \\
         \midrule
         Cross Entropy& 68.46 & 43.08 & 82.14 & \text{89.66} & \text{92.26} \\
\rowcolor{LightCyan}  Round& 65.03 $\pm$ {\small 3.59 }  & \text{47.97}$\pm$ {\small 0.94 }   & \text{86.67}$\pm$ {\small 0.20}   & 81.81$\pm$ {\small 1.56}   & 92.08$\pm$ {\small 0.42}   \\
\rowcolor{LightRed} Uniform Weighted& 71.53$\pm$ {\small 1.86}   & 46.58$\pm$ {\small0.69 }   & 86.55$\pm$ {\small 0.14}   & 84.03$\pm$ {\small 1.04}   & 92.05$\pm$ {\small 0.08}   \\\hdashline
\rowcolor{LightCyan}  DORB \cite{pasunuru-etal-2020-dorb}& 66.39$\pm$ {\small 1.84}   & 47.18$\pm$ {\small 0.41}   & 86.59$\pm$ {\small 0.25}   & 83.38$\pm$ {\small 0.21}   & 92.19$\pm$ {\small 0.21}   \\
\rowcolor{LightRed}  {\sc DynaOpt}& \text{73.80}$\pm$ {\small 1.43}   & 46.11$\pm$ {\small 0.42 }   & 86.27$\pm$ {\small 0.31}   & 85.27$\pm$ {\small 0.70}   & 91.69$\pm$ {\small 0.30}   \\
\rowcolor{LightRed}  {\sc C-DynaOpt}& 72.66$\pm$ {\small 1.29}   & 46.84$\pm$ {\small 0.67}   & 86.27$\pm$ {\small 0.21}   & 84.50$\pm$ {\small 0.81}   & 91.84$\pm$ {\small 0.38}  \\
 \bottomrule
    \end{tabular}
}
    \caption{Automated evaluation results on the counselor reflection generation task. We compute the average and standard deviation measurements of 5 different runs.
    \textcolor{cyan}{Alternate} model resuts are highlighted in \textcolor{cyan}{Cyan}.
    \textcolor{red}{Combine} model resuts are highlighted in \textcolor{red}{Red}.
    }
    \label{tab:reflection_automated_absolute}
\end{table*}

\end{document}